\pdfoutput=1

\documentclass[11pt]{article}

\usepackage[final]{acl}

\usepackage{times}
\usepackage{latexsym}

\usepackage[T1]{fontenc}

\usepackage[utf8]{inputenc}

\usepackage{microtype}

\usepackage{inconsolata}

\usepackage{graphicx}
\usepackage{enumitem}
\usepackage{url}

\usepackage{multirow}
\usepackage{wrapfig}
\usepackage{booktabs}
\usepackage{graphicx}
\usepackage{pifont}
\usepackage{arydshln}
\usepackage{color}
\usepackage{array}
\usepackage{subfigure}
\usepackage{makecell}
\usepackage{arydshln}
\usepackage{caption}
\usepackage{pbox}
\usepackage{xcolor,colortbl}
\usepackage{enumitem}
\usepackage{tcolorbox}
\usepackage{mdframed}
\usepackage{float}
\usepackage{amssymb}
\usepackage{amsmath}
\definecolor{deepgreen}{RGB}{0, 120, 0}
\definecolor{deepred}{RGB}{160, 0, 0}
\usepackage{soul}
\newcommand*{\escape}[1]{\texttt{\textbackslash#1}}
\usepackage{longtable}
\usepackage{subcaption}
\usepackage{algorithm}
\usepackage{algpseudocode}
\definecolor{deepred_code}{RGB}{0, 70, 0}
\definecolor{deepgreen_code}{RGB}{255, 0, 0}
\definecolor{brown}{rgb}{0.65, 0.16, 0.16}
\definecolor{backgreen}{RGB}{226, 240, 217}

\usepackage{listings}
\usepackage{soul}
\lstdefinestyle{python}{
    language=Python,
    basicstyle=\ttfamily\small,
    keywordstyle=\color{blue}\bfseries,
    commentstyle=\color{deepgreen_code},
    stringstyle=\color{red},
    numberstyle=\tiny\color{gray},
    showstringspaces=false,
    frame=single,
    breaklines=true,
    backgroundcolor=\color{lightgray!20}
}
\usepackage{colortbl}
\definecolor{tablegray}{gray}{.9}
\title{Curiosity-Driven Reinforcement Learning from Human Feedback}

 \author{Haoran Sun$^{*}$\, Yekun Chai\thanks{Equal contribution and shared co-first authorship. Work done during HS's internship at Baidu. \\ \indent\hspace{0.2em} $^{\spadesuit}$Correspondence to: YC. }$^{\spadesuit}$\, Shuohuan Wang \,\\ \textbf{Yu Sun \, Hua Wu \, Haifeng Wang} \\
  Baidu Inc. \\
  \texttt{sunhaoran0402@gmail.com}\\
  \texttt{\{chaiyekun,wangshuohuan\}@baidu.com}\\ 
}

\begin{document}
\maketitle
\begin{abstract}

Reinforcement learning from human feedback (RLHF) has proven effective in aligning large language models (LLMs) with human preferences, but often at the cost of reduced output diversity. This trade-off between diversity and alignment quality remains a significant challenge. Drawing inspiration from curiosity-driven exploration in reinforcement learning, we introduce curiosity-driven RLHF (CD-RLHF), a framework that incorporates intrinsic rewards for novel states, alongside traditional sparse extrinsic rewards, to optimize both output diversity and alignment quality. We demonstrate the effectiveness of CD-RLHF through extensive experiments on a range of tasks, including text summarization and instruction following. Our approach achieves significant gains in diversity on multiple diversity-oriented metrics while maintaining alignment with human preferences comparable to standard RLHF. We make our code publicly available at \url{https://github.com/ernie-research/CD-RLHF}.

\end{abstract}

\section{Introduction}
Reinforcement learning from human feedback (RLHF)~\citep{ziegler2019fine,rlhf} is now a critical component in fine-tuning large language models (LLMs) to specific tasks like code generation~\citep{roziere2023code,ernie-code,lozhkov2024starcoder}, mathematical reasoning~\citep{uesato2022solving}, and dialogue assistance~\citep{instrutGPT,anthropic-alignment}. 

Despite the success of RLHF in producing high-performing LLMs, it often reduces the output diversity of LLMs~\citep{raft,understanding-diversity}. \citet{understanding-diversity,wu2024generative} find out that models trained with RLHF exhibit a \textit{trade-off} between alignment quality and output diversity: RLHF models have high alignment quality but with low output diversity, and vice versa. 
LLMs with decreased output diversity could potentially hinder the LLMs' effectiveness on creative and open-ended tasks, like story generation~\citep{castricato2022robust,seed-story}, data synthesis~\citep{wizardlm,isheep,llama3}, and red-teaming~\citep{perez-etal-2022-red,curiosity-red-teaming}. 

Some attempts have been made to improve the diversity or achieve a balance in this trade-off.
\citet{curiosity-red-teaming} introduce SelfBLEU and Sentence-BERT metrics into the RL training as sparse rewards, mainly towards a coverage of test cases in red-teaming. Moreover, \citet{beyond-reverse-kl} gives a solution from the view of Kullback-Leibler (KL) penalty, implementing DPO or PPO with forward KL gains output diversity, but sacrifices the alignment quality. 
Therefore, is it possible to balance this trade-off in RLHF stage, thus improving output diversity \textit{without} sacrificing alignment quality?

Curiosity-driven exploration methods have been well-studied in reinforcement learning (RL) literatures~\citep{schmidhuber1991curious,bellemare2016unifying,ostrovski2017count,pathak2017curiosity,burda2018rnd}. In these RL settings, agents (\textit{i.e.,} policy models) are encouraged to explore the ``novel'' states, which are less visited in the learning progress. The curiosity signal introduced in this framework serves as a metric to estimate the ``novelty'' of a state and is implemented as an intrinsic reward. Under this setting, agents explore the ``novel'' states with curiosity, learning skills that may used to solve the latter problems.   


Inspired by these studies, we propose curiosity-driven reinforcement learning from human feedback (CD-RLHF), a novel framework that encourages agents to explore more often with curiosity in RLHF stage. As curiosity-driven RL, curiosities are implemented as intrinsic rewards and assigned along the trajectory. Combining intrinsic rewards with the extrinsic rewards generated by the reward model helps guide the agent policy in optimizing both output diversity and alignment quality. 
In CD-RLHF, we use forward dynamics to compute the prediction errors in the state representations at each time and use these errors to estimate curiosity, following~\citet{pathak2017curiosity,burda2018rnd}. This estimation shows a good property: with the training progress, a frequently visited ``novel'' state becomes boring and is less valuable for further exploration. Using curiosity encourages the agent to make various token choices in the same state potentially leading to the improvement of output diversity. Besides, the guidance of extrinsic reward ensures that the agent is still optimized towards alignment quality.

\vspace{-0.5em}\paragraph{Contribution} To conclude, our main contributions are summarized as follows:
\begin{itemize}[leftmargin=*,itemsep=2pt,topsep=0pt,parsep=2pt]
    \item We propose curiosity-driven RLHF, a framework that maintains alignment performance comparable to RLHF while achieving higher output diversity. We demonstrate its effectiveness through extensive experiments across various datasets and tasks, including text summarization and instruction following. 
    \item We demonstrate that CD-RLHF improves output diversity evaluated by diversity-oriented metrics in both lexical and semantic dimensions, while preserving alignment quality comparable to vanilla RLHF.
    \item In the out-of-distribution benchmark MT-Bench, CD-RLHF demonstrates strong generalization capability, achieving higher MT-Bench scores compared to RLHF evaluated by GPT-4, and consistently showing more diversity outputs. 
\end{itemize}
\section{Related Work}
\paragraph{LLM Alignment} RLHF significantly enhances the alignment of LLMs, improving instruction following, safety, and reliability in real-world applications~\citep{ziegler2019fine,hhh,li2024toolaugmented}. Recent efforts in efficient alignment algorithms focus on general performance improvement~\citep{remax,ahmadian2024back,MA-RLHF}, directly aligning with preference data~\citep{dpo,cpo,ipo}, and tacking credit assignment with dense rewards~\citep{wu2023fine,attention-reward,tlcr}. However, alignment can degrade output diversity, as pointed out by \citet{understanding-diversity} who found that RLHF-trained LLMs show less diversity than supervised fine-tuned models. Nevertheless, \citet{wu2024generative} defines this as generative monoculture, which means LLMs refuse to share diverse options. There are some attempts to try to mitigate this situation with different approaches. \citet{beyond-reverse-kl} proposes using $f$-divergence with DPO/PPO to improve diversity at the cost of alignment, different divergences show a different level of trade-off between diversity and alignment. \citet{curiosity-red-teaming} integrate SelfBLEU and Sentence-BERT into RL fine-tuning, boosting output diversity for the improvement of coverage of test cases in red-teaming tasks. \citet{qdaif} proposes a method that is applied at the inference stage, generating a new solution based a previous solution. These solutions are evaluated with LLMs considering both quality and diversity, and the better one is reserved. 

\begin{figure*}[t]
    \centering
    \includegraphics[trim={0pt 10pt 0pt 0}, width=0.85\textwidth]{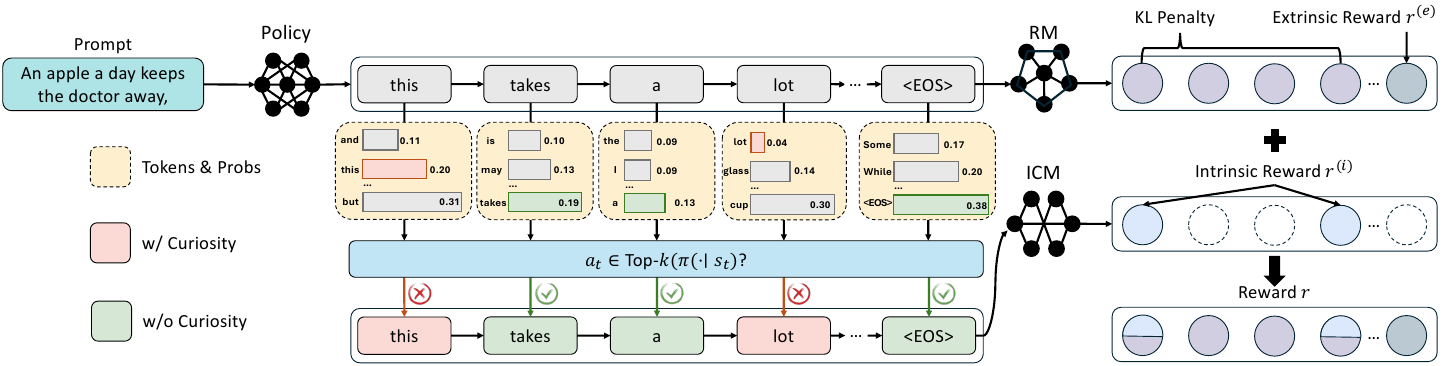}
    \caption{Illustration of the CD-RLHF framework. The policy model generates a completion based on the given instruction, which samples tokens from vocabulary at each time. The introduced intrinsic curiosity module (ICM) estimates the curiosity as a metric for ``novelty'' of the context, producing the intrinsic rewards. Another mechanism is introduced to determine which context is worth exploring, based on whether the selected token's probability ranks among the top‑$k$ highest probabilities ($k = 1$ in the illustration).}
    \label{fig:cdrlhf_framework}
\end{figure*}

\paragraph{Curiosity-Driven RL} Curiosity-driven approaches have been widely studied in RL. Count-based approaches are successful in tabular settings but fail at generalising to additional tasks with infinite states~\citep{bellemare2016unifying,ostrovski2017count}. Using prediction error as curiosity approaches highlight their success in multiple RL scenarios~\citep{schmidhuber1991curious,schmidhuber1991possibility,stadie2015incentivizing}. Specifically, methods proposed by~\cite{pathak2017curiosity,burda2018largescale,burda2018rnd} treat curiosities as intrinsic rewards, and use intrinsic rewards to encourage agents to explore at novel states, intending to enhance agent capabilities that could be used in future environments. Pure curiosity-driven learning has also been studied in large-scale settings~\citep{burda2018largescale}. 
\section{Method}

In this section, we introduce our CD-RLHF method. The key idea is to encourage the policy to explore states with high curiosity potential, which may yield a broader range of diverse tokens and thus improve output diversity.

\subsection{Preliminaries: Curiosity-Driven RL}
Curiosity-driven RL~\citep{pathak2017curiosity,burda2018rnd}, also known as exploration bonus, encourages an agent to explore novel states with curiosities that serve as the intrinsic rewards $r^{(i)}$. By doing so, the sparse extrinsic reward $r^{(e)}$ which guides the agent from a holistic perspective is replaced with a new reward $r_t = r^{(i)}_t+r^{(e)}_t$. The $r^{(e)}_t$ is mostly near zero along the trajectory, except for the end of the episode. 

The motivation behind curiosity-driven RL is that intrinsic rewards should be higher in novel states than in normal states to encourage the agent to visit. 
A common approach in the RL literature is to estimate intrinsic rewards (\textit{i.e.}, curiosity) using prediction errors. The forward dynamics~\cite{schmidhuber1991curious,pathak2017curiosity,burda2018rnd} or inverse dynamics~\cite{haber2018learning} are utilized to predict the representation of the next state, then compute the prediction error between the predicted and ground truth representation of the next state.
By maximizing the prediction errors in RL procedure, agents tend to get attracted to the stochastic transitions, which correspond to novel states. During the training progress, the prediction errors tend to decrease as the agent becomes familiar with more states. Since the prediction happens in the latent space, the choice of features to estimate the prediction error is a vital factor. As indicated by~\citet{burda2018largescale}, the latent variables generalise better than the raw observations (\textit{i.e.}, pixel), making them more effective for estimating the intrinsic rewards. 

\subsection{Curiosity-Driven Alignment} \label{sec:review}


Our insight is that the curiosity in RLHF reflects the agent's unfamiliarity with the state. By frequently exploring such state with high curiosity, the agent is encouraged to select diverse actions under the same state across trajectories, thereby enhancing output diversity. In CD-RLHF, we integrate a intrinsic curiosity module (ICM) into RLHF framework to estimate this curiosity. The estimation happens only at the states with enough ``novelty'', which is determined by the probability of the selected token, as shown in Figure~\ref{fig:cdrlhf_framework}.

\subsubsection{Curiosity-Driven Reward}
The extrinsic reward $r^{(e)}_t$ at time step $t$ is produced by an external reward model, and obtained at the end of the episode. In RLHF, the Kullback-Leibler (KL) penalty works as a token-level reward, which constrains the policy’s divergence from the supervised fine-tuned (reference) model:
\begin{equation*}
    r^{(e)} = R - \beta D_{\mathrm{KL}}(\pi_\mathrm{policy}(\cdot) ~\|~ \pi_{\mathrm{ref}}(\cdot))
\end{equation*}
where $R$ is the reward produced by the reward model, $\pi_\mathrm{policy}$ is the policy model and $\pi_\mathrm{ref}$ is the reference model.

Combining intrinsic reward and extrinsic reward, we can obtain the reward that is used to optimize the policy in CD-RLHF:
\begin{equation*}
    r_t = r^{(e)}_t + \eta \cdot r^{(i)}_t
\end{equation*}
where the discounting factor $\eta$ controls the scale of intrinsic rewards.

This reward is designed to steer policy optimization toward achieving both alignment quality, driven by extrinsic rewards, and output diversity, driven by intrinsic rewards.

\subsubsection{Prediction Error as Curiosity Reward} \label{sec:estimate}
In the RLHF framework, the state consists of a sequence of action tokens up to current time step, and the action selected at each step represents the decision made. At time step $t$, the action is denoted as $a_t$, and the state is represented as $s_t = \{s_0, a_{<t}\}$, where $s_0$ is the initial prompt and $a_{<t}$ refers to the sequence of actions taken before time $t$. The transition function can be interpreted as updating the state with the generated token at each time step, where the state at time $t+1$ is represented as $s_{t+1} = \{s_t, a_t\} $, reflecting the concatenation of the previous state $s_t$ and the newly generated token $a_t$.


\paragraph{Intrinsic Curiosity Module (ICM)} 
To estimate intrinsic rewards, we introduce an Intrinsic Curiosity Module (ICM), inspired by \citet{pathak2017curiosity, burda2018rnd}. The ICM predicts the next state representation based on the current state and action, and the prediction error serves as the curiosity-driven intrinsic reward. This prediction error is computed in the latent space of hidden embeddings from large language models (LLMs), enabling a more expressive and informative representation of states and actions.




ICM consists of a feature encoder $\phi$ and a forward model $f$, both implemented as two-layer MLPs. The feature encoder $\phi$ encodes the state representation to bridge the scalar gap between raw state and action representations. The forward model $f$ takes as input the concatenation of the encoded state $\phi(s_t)$ and the action representation $\psi(a_t)$, predicting the next state representation $\phi(s_{t+1})$. This prediction process is formalized as:
\begin{equation}
    \hat \phi(s_{t+1}) = f(\phi(s_{t}), \psi(a_t)) 
\end{equation}

The ICM loss $\mathcal{L}_{\mathrm{ICM}}$ is computed as the squared error between the predicted and actual next-state representations:
\begin{equation}
    \mathcal{L}_{\mathrm{ICM}} = \frac{1}{2}\|\hat \phi(s_{t+1}) - \phi(s_{t+1})\|_2^2
\end{equation}
where $\hat \phi(s_{t+1})$ is the predicted state representation, $\phi(s_{t+1})$ is the true latent representation. 



\paragraph{Intrinsic Reward Computation} The ICM is trained independently of the policy model, optimizing $\mathcal{L}_{\mathrm{ICM}}$ as a self-supervised learning task. The resulting prediction error is used as the intrinsic reward, encouraging the exploration of less predictable states.  Intuitively, deterministic states with limited action choices offer little exploratory value. To improve the efficiency, we apply a top-$k$ strategy to determine whether a state warrants further exploration. The intrinsic reward $r^{(i)}_t$ at time step $t+1$ is defined as:

\vspace{-0.1 em}
\begin{equation*}
r^{(i)}_t=\begin{cases}
         0 \text{ if } a_t \in \text{Top-}k(\pi(\cdot\mid s_t)) \\ \frac{1}{2} \|\hat \phi(s_{t+1}) - \phi(s_{t+1})\|_2 \text{ otherwise }
    \end{cases}
\end{equation*}
Here, $\text{Top-}k(\pi(\cdot\mid s_t))$ denotes the top-$k$ vocabulary subset that maximizes $\sum_{a_t\in \text{Top-}k(\cdot)}\pi\left( a_t \mid s_{t} \right)$. When the selected action $a_t$ is within the top-$k$ most probable candidates, the intrinsic reward is set to zero to discourage redundant exploration.



\begin{figure}[t]
    \centering
    \includegraphics[trim={0pt 10pt 0pt 0}, width=0.38\textwidth]{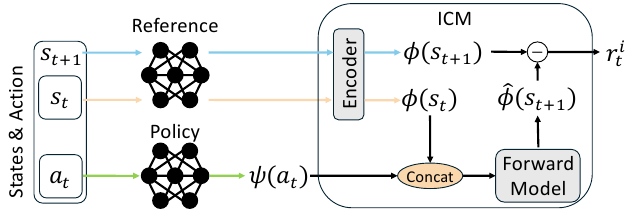}
    \caption{
    The computation of ICM loss $\mathcal{L}_{\rm{ICM}}$ and intrinsic reward $r_t^{(i)}$ at time $t$, the state representation $s_t$ and $s_{t+1}$ are derived from the reference model, while the action representation $a_t$ is taken from the policy model.
    }
    \label{fig:cdrlhf_compute}
    \vspace{-1em}
\end{figure}

The intrinsic reward computation aligns with $\mathcal{L}_{\mathrm{ICM}}$, as the state prediction accuracy improves during ICM training, reducing the intrinsic reward over time. This follows the principle of curiosity-driven RL—novel states become less rewarding with frequent visits.  Figure~\ref{fig:cdrlhf_compute} illustrates the intrinsic reward estimation process.


\paragraph{Reward Whitening} The scalar gap between intrinsic and extrinsic rewards could introduce unstable optimization in RL training, where intrinsic rewards may vary across samples. To mitigate this, we apply a \textit{reward whitening} to normalize intrinsic rewards at each time step along the trajectory:
\begin{equation*}
    r^{(i)} = (r^{(i)} - \mu) / \sigma^2
\end{equation*}
where $\mu$ and $\sigma$ represent the mean and standard deviation of the intrinsic rewards, respectively.

\paragraph{Feature Space Alignment} During the intrinsic reward estimation in RLHF, there is a gap between predictions and the states/actions. Predictions are operated in continuous space, while states and actions reside in discrete space. To tackle this gap, we use the last layers' hidden states $s_t$ derived from the reference model, and use the embeddings of the actor model to extract the representation of $a_t$. This alignment results in a coherent feature space for computing prediction errors.

\section{Experiments}

\subsection{Experimental Settings} \label{sec:exp_setting}

\paragraph{Datasets} We evaluate our method on the TL;DR~\citep{summarize_feedback} dataset for text summarization, which includes 93k human-annotated preference pairs and 86k pairs for evaluation. Additionally, we evaluate on the UltraFeedback~\citep{UltraFeedback} dataset for instruction following, which contains 61.1k pairs for training and 2k pairs for evaluation.\footnote{We use the binarized version from \url{https://huggingface.co/datasets/HuggingFaceH4/UltraFeedback_binarized}, following the preference modeling splits in all experiments.} 

\paragraph{Models and training details} We use pre-trained Gemma-2B, Gemma-7B~\citep{gemma}, Llama-3.2-1B, and Llama-3.2-3B~\citep{llama3} as base models. CD-RLHF is implemented using the DeepSpeed-Chat~\citep{deepspeed-chat} framework. The datasets are split into SFT/RM/PPO partitions with a 20\%/40\%/40\% ratio, respectively. Hyperparameters used for SFT, RM, and PPO stages are detailed in Appendix~\ref{app:training_details}. In CD-RLHF, we use top-$1$ to select valid intrinsic rewards.

\begin{table*}[t]
\centering
\scalebox{0.58}{
\begin{tabular}{l|l|cccc|c}
\toprule
\textbf{Stage}                 & \textbf{Model}                                                                                 & \textbf{Diversity $\uparrow$}                                                  & \textbf{EAD $\uparrow$}                                                       & \textbf{SelfBLEU $\downarrow$}                                                 & \textbf{SentBERT $\downarrow$}                                                & \textbf{RM Score $\uparrow$}                   \\
\midrule
\multicolumn{7}{c}{\textbf{TL;DR}}                                                                                                                                                                                                                                                                                                                                                                                                                       \\
\midrule
\rowcolor{tablegray}
                    SFT &                                                                                     & 0.4041                                                               & 0.8444                                                               & 0.1336                                                               & 0.5380                                                                                     & -0.52                                \\
                    RLHF &                                                                                     & 0.2132                                                               & 0.7347                                                               & 0.3367                                                               & 0.7024                                                                                     & 0.90                                 \\
                    Sent-Rewards~\citep{curiosity-red-teaming}  &                                                                   & \ul{0.2355}                                                               & \ul{0.7512}                                                               & \ul{0.3053}                                                               & \ul{0.6961}                                                                                     & \textbf{0.95}                        \\
\begin{tabular}[c]{@{}l@{}}CD-RLHF\\ ($\Delta$ over RLHF)\end{tabular} & \multirow{-4}{*}{Gemma-2B}  & \textbf{\begin{tabular}[c]{@{}c@{}}0.2839\\ (\textcolor{deepgreen}{+33.16\%})\end{tabular}} & \textbf{\begin{tabular}[c]{@{}c@{}}0.7793\\ (\textcolor{deepgreen}{+6.07\%})\end{tabular}}  & \textbf{\begin{tabular}[c]{@{}c@{}}0.2590\\ (\textcolor{deepgreen}{+23.08\%})\end{tabular}} & \textbf{\begin{tabular}[c]{@{}c@{}}0.6720\\ (\textcolor{deepgreen}{+4.33\%})\end{tabular}}                        & \textbf{\begin{tabular}[c]{@{}c@{}}0.95\\ \ \end{tabular}} \\
\midrule
\rowcolor{tablegray}
                    SFT &                                                                                     & 0.3575                                                               & 0.8256                                                               & 0.1706                                                               & 0.5442                                                                                     & {0.62}          \\
                    RLHF &                                                                                     & 0.1180                                                               & 0.6602                                                               & 0.4352                                                               & \ul{0.7601}                                                                                     & {\textbf{2.02}} \\
                    Sent-Rewards~\cite{curiosity-red-teaming} & & \textbf{0.1447} & \ul{0.6735} & \ul{0.4197} & 0.7749 & \ul{1.87} \\
\begin{tabular}[c]{@{}l@{}}CD-RLHF\\ ($\Delta$ over RLHF)\end{tabular} & \multirow{-3}{*}{Gemma-7B} & {\begin{tabular}[c]{@{}c@{}}\ul{0.1360}\\ (\textcolor{deepgreen}{\textbf{+15.25\%}})\end{tabular}} & \textbf{\begin{tabular}[c]{@{}c@{}}0.6816\\ (\textcolor{deepgreen}{+3.24\%})\end{tabular}}  & \textbf{\begin{tabular}[c]{@{}c@{}}0.4144\\ (\textcolor{deepgreen}{+4.78\%})\end{tabular}}  & {\textbf{\begin{tabular}[c]{@{}c@{}}0.7480\\ (\textcolor{deepgreen}{+1.59\%})\end{tabular}}} & {\begin{tabular}[c]{@{}c@{}}\textbf{2.02}\\ \ \end{tabular}} \\
\midrule
\rowcolor{tablegray}
                                        SFT &                                                                                     & 0.3672                                                               & 0.8296                                                               & 0.1651                                                               & 0.5498                                                                                     & -0.31                                \\
                    RLHF &                                                                                     & 0.1724                                                               & 0.6869                                                               & 0.3997                                                               & \ul{0.6971}                                                                                     & \ul{1.14} \\
                    Sent-Rewards~\cite{curiosity-red-teaming} & & \ul{0.1869} & \ul{0.7078} & \ul{0.3519} & 0.7338 & 0.80 \\
\begin{tabular}[c]{@{}l@{}}CD-RLHF\\ ($\Delta$ over RLHF)\end{tabular} & \multirow{-3}{*}{Llama-3.2-1B} & \textbf{\begin{tabular}[c]{@{}c@{}}0.2418\\ (\textcolor{deepgreen}{+40.26\%})\end{tabular}} & \textbf{\begin{tabular}[c]{@{}c@{}}0.7482\\ (\textcolor{deepgreen}{+8.92\%})\end{tabular}}  & \textbf{\begin{tabular}[c]{@{}c@{}}0.3108\\ (\textcolor{deepgreen}{+22.24\%})\end{tabular}}  & {\textbf{\begin{tabular}[c]{@{}c@{}}0.6847\\ (\textcolor{deepgreen}{+1.78\%})\end{tabular}}} & {\begin{tabular}[c]{@{}c@{}}\textbf{1.17}\\ \ \end{tabular}} \\
\midrule
\rowcolor{tablegray}
                                        SFT &                                                                                     & 0.3539                                                               & 0.8233                                                               & 0.1753                                                               & 0.5649                                                                                     & 2.32                                \\
                    RLHF &                                                                                     & 0.2281                                                               & 0.7441                                                               & \ul{0.3163}                                                               & \ul{0.6658}                                                                                     & 3.33 \\
                    Sent-Rewards~\citep{curiosity-red-teaming} & & \ul{0.2355} & \ul{0.7530} & 0.3356 & 0.7182 & \textbf{3.54} \\
\begin{tabular}[c]{@{}l@{}}CD-RLHF\\ ($\Delta$ over RLHF)\end{tabular} & \multirow{-4}{*}{Llama-3.2-3B} & \textbf{\begin{tabular}[c]{@{}c@{}}0.2920\\ (\textcolor{deepgreen}{+28.01\%})\end{tabular}} & \textbf{\begin{tabular}[c]{@{}c@{}}0.7879\\ (\textcolor{deepgreen}{+5.89\%})\end{tabular}}  & \textbf{\begin{tabular}[c]{@{}c@{}}0.2463\\ (\textcolor{deepgreen}{+22.13\%})\end{tabular}}  & {\textbf{\begin{tabular}[c]{@{}c@{}}0.6551\\ (\textcolor{deepgreen}{+1.61\%})\end{tabular}}} & {\begin{tabular}[c]{@{}c@{}}\ul{3.49}\\ \ \end{tabular}} \\
\midrule
\multicolumn{7}{c}{\textbf{UltraFeedback}}                  \\
\midrule
\rowcolor{tablegray}
                                        SFT &                                                                                     & 0.1855                                                               & 0.7346                                                               & 0.2899                                                               & 0.6766                                                                                     & -1.45                                \\
                                        RLHF &                                                                                     & \ul{0.1686}                                                               & 0.6503                                                               & \ul{0.3104}                                                               & 0.7672                                                                                     & -1.01                                \\
                                       Sent-Rewards~\citep{curiosity-red-teaming}   &  & 0.1603                                                               & \ul{0.6801}                                                               & 0.3483                                                               & \ul{0.7463}                                                                                     & \textbf{-0.80}                       \\
\begin{tabular}[c]{@{}l@{}}CD-RLHF\\ ($\Delta$ over RLHF)\end{tabular}   & \multirow{-4}{*}{Gemma-2B} & \textbf{\begin{tabular}[c]{@{}c@{}}0.1899\\ (\textcolor{deepgreen}{+12.63\%})\end{tabular}} & \textbf{\begin{tabular}[c]{@{}c@{}}0.7417\\ (\textcolor{deepgreen}{+14.06\%})\end{tabular}} & \textbf{\begin{tabular}[c]{@{}c@{}}0.2858\\ (\textcolor{deepgreen}{+7.93\%})\end{tabular}}  & \textbf{\begin{tabular}[c]{@{}c@{}}0.7308\\ (\textcolor{deepgreen}{+4.74\%})\end{tabular}}                        & \begin{tabular}[c]{@{}c@{}}\ul{-0.90}\\ \ \end{tabular}         \\
\midrule
\rowcolor{tablegray}
                                        SFT &                                                                                     & 0.2148                                                               & 0.7594                                                               & 0.2292                                                               & 0.6378                                                                                     & 0.29                                 \\
                                        RLHF &                                                                                     & 0.2345                                                               & 0.7360                                                               & 0.2717                                                               & 0.7298                                                                                     & \textbf{0.63}                        \\
                                        Sent-Rewards~\cite{curiosity-red-teaming} & & \ul{0.2557} & \textbf{0.7744} & \ul{0.2512} &  \ul{0.7231} & 0.60 \\
\begin{tabular}[c]{@{}l@{}}CD-RLHF\\ ($\Delta$ over RLHF)\end{tabular} & \multirow{-3}{*}{Gemma-7B} & \textbf{\begin{tabular}[c]{@{}c@{}}0.2654\\ (\textcolor{deepgreen}{+13.18\%})\end{tabular}} & {\begin{tabular}[c]{@{}c@{}}\ul{0.7639}\\ (\textcolor{deepgreen}{\textbf{+3.79\%}})\end{tabular}}  & \textbf{\begin{tabular}[c]{@{}c@{}}0.2442\\ (\textcolor{deepgreen}{+10.12\%})\end{tabular}} & \textbf{\begin{tabular}[c]{@{}c@{}}0.6858\\ (\textcolor{deepgreen}{+6.03\%})\end{tabular}}                        & \begin{tabular}[c]{@{}c@{}}\ul{0.62}\\ \ \end{tabular}                               \\
\midrule
\rowcolor{tablegray}
                                        SFT &                                                                                     & 0.2060                                                               & 0.7545                                                               & 0.2476                                                               & 0.6517                                                                                     & -0.03                                \\
                    RLHF &                                                                                     & \ul{0.1683}                                                               & \ul{0.6499}                                                               & 0.3564                                                               & \ul{0.7813}                                                                                     & \textbf{1.00} \\
                    Sent-Rewards~\cite{curiosity-red-teaming} & & 0.1652 & 0.6170 & \ul{0.3345} & 0.7924 & \ul{0.97} \\
\begin{tabular}[c]{@{}l@{}}CD-RLHF\\ ($\Delta$ over RLHF)\end{tabular} & \multirow{-3}{*}{Llama-3.2-1B}  & \textbf{\begin{tabular}[c]{@{}c@{}}0.1834\\ (\textcolor{deepgreen}{+8.97\%})\end{tabular}} & \textbf{\begin{tabular}[c]{@{}c@{}}0.6891\\ (\textcolor{deepgreen}{+6.03\%})\end{tabular}}  & \textbf{\begin{tabular}[c]{@{}c@{}}0.3149\\ (\textcolor{deepgreen}{+11.64\%})\end{tabular}}  & {\textbf{\begin{tabular}[c]{@{}c@{}}0.7598\\ (\textcolor{deepgreen}{+2.75\%})\end{tabular}}} & \begin{tabular}[c]{@{}c@{}}\ul{0.97}\\ \ \end{tabular} \\
\midrule
\rowcolor{tablegray}
                                        SFT &                                                                                     & 0.1974                                                               & 0.7484                                                               & 0.2669                                                               & 0.6885                                                                                     & 0.59                                \\
                    RLHF &                                                                                     & \ul{0.1805}                                                               & 0.7031                                                               & 0.3188                                                               & 0.7676                                                                                     & \ul{1.35} \\
                    Sent-Rewards~\cite{curiosity-red-teaming} & & 0.1765 & \ul{0.7261} & \ul{0.2944} & \ul{0.7471} & 1.22 \\
\begin{tabular}[c]{@{}l@{}}CD-RLHF\\ ($\Delta$ over RLHF)\end{tabular} & \multirow{-3}{*}{Llama-3.2-3B}  & \textbf{\begin{tabular}[c]{@{}c@{}}0.2223\\ (\textcolor{deepgreen}{+23.16\%})\end{tabular}} & \textbf{\begin{tabular}[c]{@{}c@{}}0.7673\\ (\textcolor{deepgreen}{+9.13\%})\end{tabular}}  & \textbf{\begin{tabular}[c]{@{}c@{}}0.2531\\ (\textcolor{deepgreen}{+20.61\%})\end{tabular}}  & {\textbf{\begin{tabular}[c]{@{}c@{}}0.7349\\ (\textcolor{deepgreen}{+4.26\%})\end{tabular}}} & {\textbf{\begin{tabular}[c]{@{}c@{}}1.43\\ \ \end{tabular}}} \\
\bottomrule
\end{tabular}
}
\caption{Results on TL;DR and UltraFeedback datasets, comparing CD-RLHF with SFT, RLHF, and Sent-Rewards~\citep{curiosity-red-teaming} from both output diversity and alignment quality aspects. }
\label{tab:main_results}
\vspace{-1.5em}
\end{table*}

\paragraph{Compared baselines} In addition to the vanilla RLHF baseline, we evaluate other approaches aimed at enhancing output diversity. \citet{curiosity-red-teaming} propose optimizing SelfBLEU and Sentence-BERT scores across samples during RLHF training as sentence-level additional rewards, and optimizing the entropy as token-level additional rewards. We denote this method as Sent-Rewards and re-implement it within DeepSpeed-Chat, using consistent hyperparameters in PPO training for fair comparison. Additional details of implementing Sent-Rewards can be found in Appendix~\ref{app:training_details}.

\paragraph{Evaluation} We evaluate performance using both quality and diversity metrics. 
\begin{itemize}[leftmargin=*,itemsep=2pt,topsep=0pt,parsep=2pt]
    \item \textbf{RM scores}: RM scores evaluate the alignment quality using the trained reward model, which is identical to the one used in PPO training.
    \item \textbf{$N$-gram distinct}~\citep{li2016diversity}: It measures the number of distinct $N$-grams in total $N$-grams (averaging 1- to 5-grams). This metric is referred as Diversity. 
    \item \textbf{EAD}~\citep{ead}: EAD is a refined version of $N$-gram Distinct, EAD adjust for length bias by introducing vocabulary size.
    \item \textbf{SelfBLEU}~\citep{selfbleu}: It calculates BLEU scores within outputs, by taking one sentence as hypothesis and the rest as reference to assess lexical diversity (averaging 1- to 4-grams).
    \item \textbf{Sentence-BERT}~\citep{sentbert}: It computes semantic diversity by averaging cosine similarity between output embeddings. We refer to this metric as SentBERT. 
    \item \textbf{GPT-4 win-rate}~\citep{tevet-berant-2021-evaluating}: We utilize GPT-4 to evaluate the diversity by choosing the more diversity set from two sets of completions, which are produced by different models. 
    \item \textbf{Human win-rate}: Similar to GPT-4 win-rate, we assess the win-rates between two methods with the assistance of human annotators. 
\end{itemize}
For the automated metrics, we randomly sample 2k instances from each dataset’s validation/test set, generating 10 completions for an instruction with a temperature of 1.0. For diversity related metrics, we compute the per-input diversity which estimate the diversity within the 10 completions for an instruction, and averaged across instances. We sampled 50 instances from the 2k subset to assess the win-rate estimated by GPT-4. To decrease the evaluation difficulty of human evaluation, we sampled 25 instances from GPT-4 evaluation subset, with each instruction has 5 completions. 


\subsection{Main Results} \label{sec:exp_main_results}
The evaluation results are shown in Table~\ref{tab:main_results}. Overall, CD-RLHF consistently outperforms the PPO baseline and Sent-Rewards in terms of output diversity, while achieving a favorable balance between output diversity and alignment quality.

\paragraph{Results on TL;DR} For the TL;DR summarization task, both CD-RLHF and Sent-Rewards improve output diversity over the PPO baseline. CD-RLHF achieving the highest diversity scores among all RLHF models, obtains 16.66\% improvements on Gemma-2B and 6.22\% on Gemma-7B compared to RLHF baseline. Additionally, CD-RLHF maintains comparable alignment quality (RM scores) to the PPO baseline. For Llama-3.2-1B and Llama-3.2-3B, CD-RLHF achieves remarkable performance gain on both output diversity, with 18.30\% and 14.41\% improvements on output diversity in average, respectively. Besides, CD-RLHF on Llama-3.2-3B brings improvements on alignment quality with +0.16 in RM scores.

\paragraph{Results on UltraFeedback} For the UltraFeedback instruction-following task, CD-RLHF demonstrates substantial improvements in output diversity for all models. While the Sent-Rewards method helps mitigate output diversity degradation, it remains less effective than CD-RLHF in balancing this trade-off. In this setting, CD-RLHF effectively manages the trade-off, showing only a minimal drop in quality on the Gemma-7B (-0.01 in RM score) and Llama-3.2-1B (-0.03 in RM score). Compared to RLHF baseline, CD-RLHF achieves an average of 9.84\% on Gemma-2B, 8.28\% on Gemma-7B, 7.35\% on Llama-3.2-1B, and 14.29\% on Llama-3.2-3B. Additionally, CD-RLHF outperforms the SFT model in lexical diversity, achieving higher scores on Diversity, EAD, and SelfBLEU metrics for the Gemma-2B and Llama-3.2-3B, and on Diversity and EAD metrics for the Gemma-7B.

\begin{figure}[t] 
	\centering
	\subfigure[GPT-4 Win-Rates]{
	    \centering
            \label{fig:win_rate_rnkGDS}
		\includegraphics[trim={0pt 10pt 0pt 0}, width=0.9\linewidth]{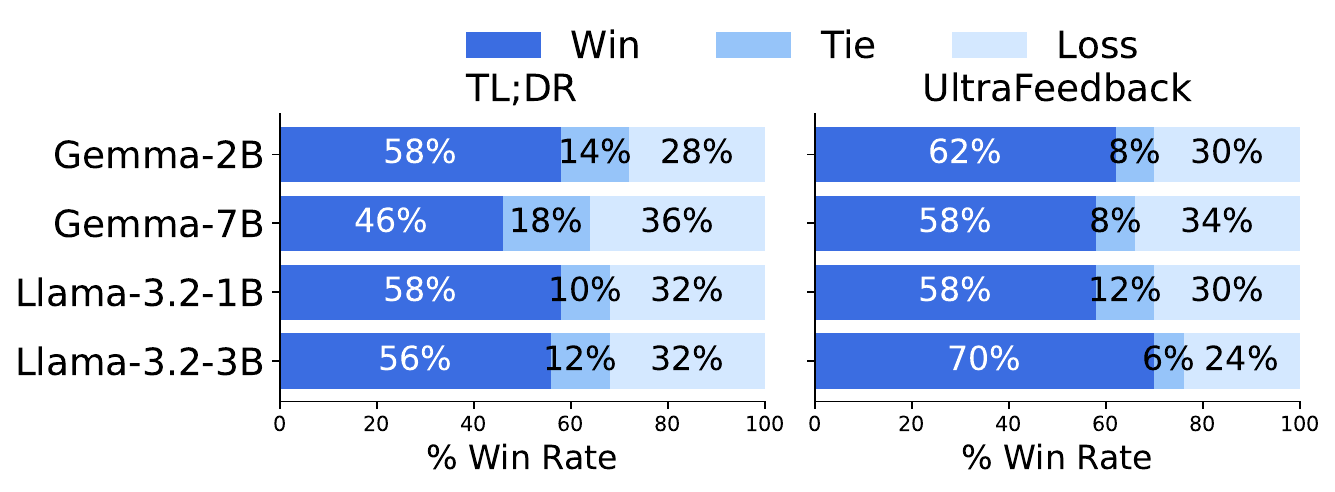}
        }\\
        \vspace{-1em}
	\subfigure[Human Win-Rates]{
	    \centering
            \label{fig:win_rate_human}
		\includegraphics[trim={0pt 10pt 0pt 0}, width=0.9\linewidth]{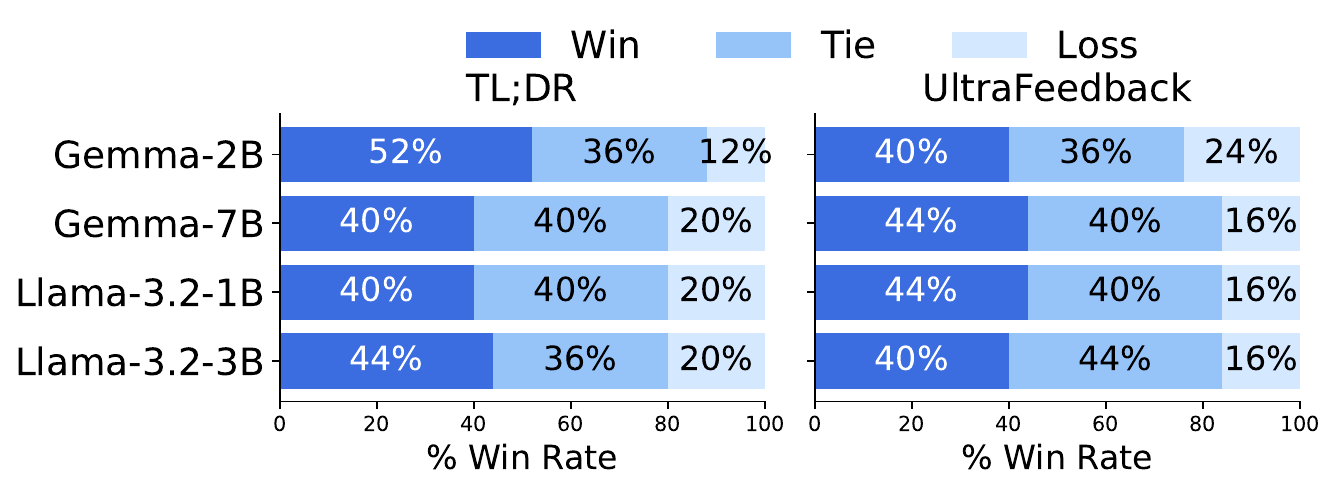}
        }%
        \vspace{-1em}
	\caption{The win rates of CD-RLHF against vanilla PPO when evaluating output diversity on TL;DR and UltraFeedback datasets.}
        \vspace{-1em}
\end{figure}

\begin{figure}[t]
    \centering
    \includegraphics[trim={0pt 30pt 0pt 0}, width=0.44\textwidth]{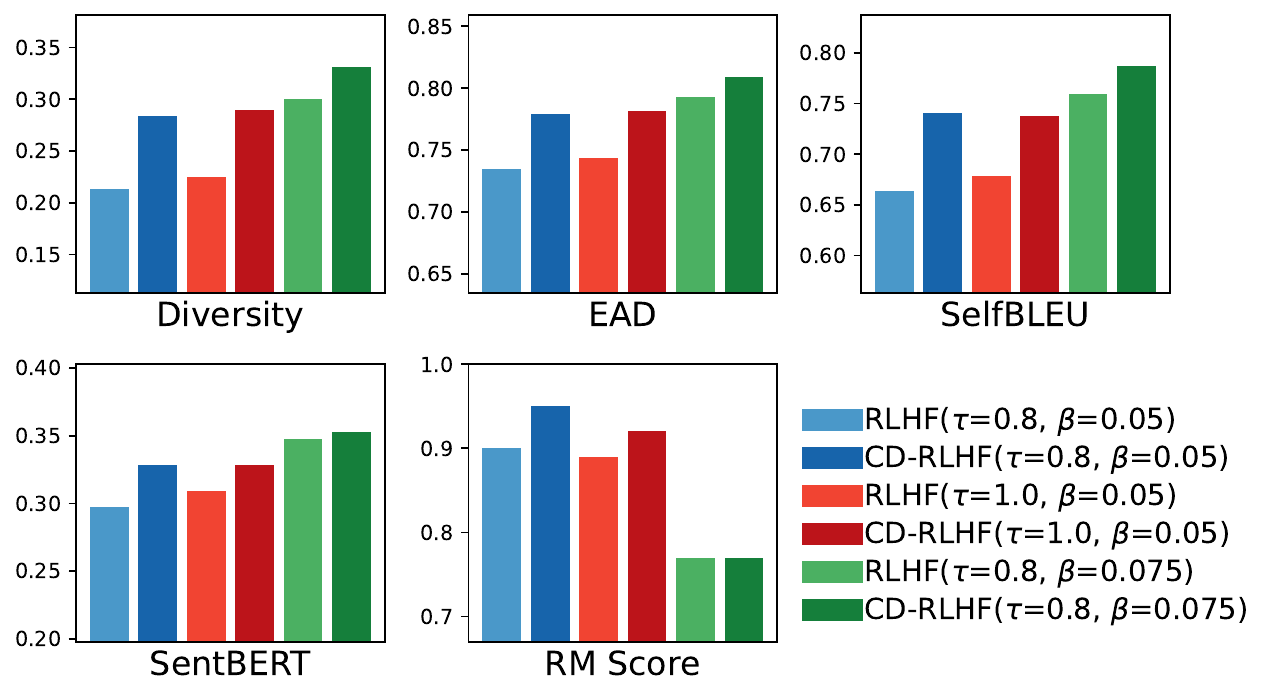}
    \caption{Output diversity and alignment performance of RLHF and CD-RLHF under different hyperparameter settings: KL coefficient $\beta$ and sampling temperature $\tau$. For a better illustration, we use 1.0 - SelfBLEU and 1.0 - SentBERT as metric to maintain a similar trend with other metrics.}
    \label{fig:kl}
    \vspace{-1.5em}
\end{figure}
\paragraph{GPT-4 and human evaluation} For further evaluation using GPT-4, as shown in Figure~\ref{fig:win_rate_rnkGDS}, CD-RLHF achieves a win rate of 58\%, 46\%, 58\% and 56\% over the PPO baseline on the TL;DR dataset for the Gemma-2B, Gemma-7B, Llama-3.2-1B and Llama-3.2-3B, respectively. On the UltraFeedback dataset, CD-RLHF consistently yields an average 62\% win rates across the evaluated models. In Figure~\ref{fig:win_rate_human}, we demonstrate the human evaluations on the TL;DR and UltraFeedback datasets, where CD-RLHF consistently outperforms RLHF on all evaluated models. These comparison evaluation results demonstrate the remarkable performance CD-RLHF achieved in improving output diversity.

\subsection{Ablation Studies} \label{sec:exp_ablation}
In this section, we conduct ablation studies on Gemma-2B for better analysis with various hyper-parameter settings.

As previous works~\citep{understanding-diversity,beyond-reverse-kl}  have shown, the KL divergence in RLHF heavily influences the balance between output diversity and alignment performance. In Figure~\ref{fig:kl} (blue and red bars), we investigate CD-RLHF's performance under various KL settings: KL coefficient $\beta=0.05$ and $0.075$. Within each method, we observe that as $\beta$ increases, output diversity improves, albeit at the cost of alignment performance. However, under the same KL setting, CD-RLHF consistently outperforms RLHF in output diversity, without compromising alignment quality.

Another factor that could affect the performance of CD-RLHF is the choice of the sampling hyperparameters. In this section, we examine the effect of sampling with different settings: temperature $\tau=0.8$ and $1.0$. Figure~\ref{fig:kl} (blue and green bars) shows the performance of RLHF and CD-RLHF under these sampling temperatures during training. While increasing the sampling temperature during experience generation moderately improves output diversity, its impact is less pronounced than that of increasing the KL. Under both temperature settings, CD-RLHF still show improvements on output diversity, and also the alignment quality.

\begin{figure}[t]
    \centering
    \includegraphics[trim={0pt 30pt 0pt 0}, width=0.4\textwidth]{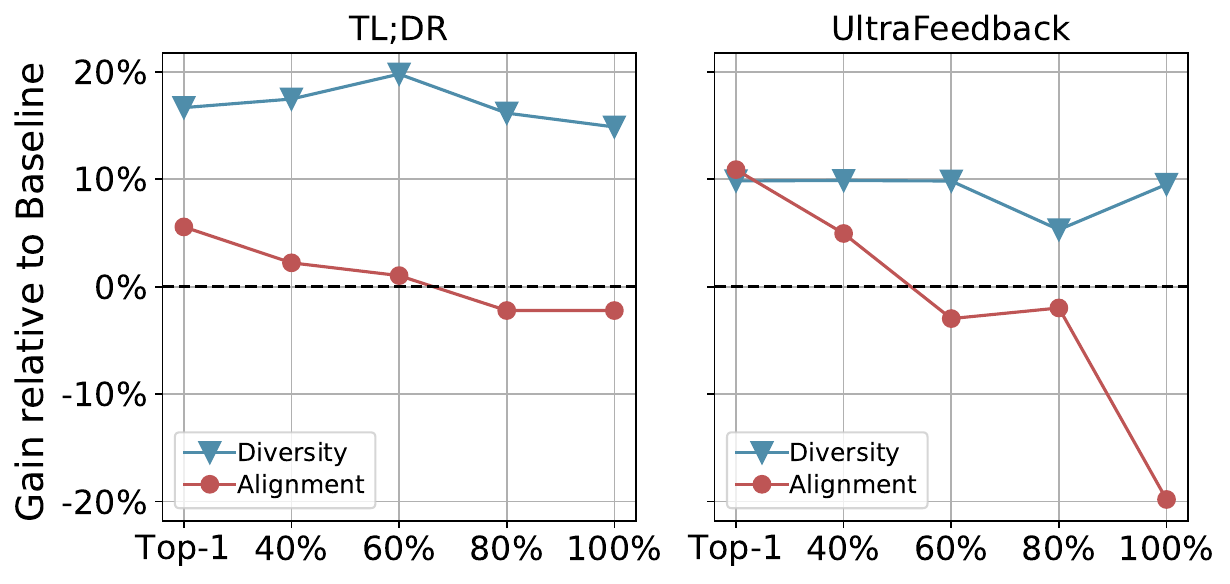}
    \caption{Diversity and alignment quality changes with increasing the frequency of intrinsic rewards. The y-axis represents the improvement ratio against RLHF baseline. The diversity is the average of the related metrics: Diversity, EAD, SelfBLEU, and SentBERT, and alignment is the RM score.}
    \vspace{-0.5em}
    \label{fig:ratio}
    \vspace{-1em}
\end{figure}
\subsection{Analysis} \label{sec:identify}
\paragraph{Frequency of intrinsic rewards} In the computation described in \S\ref{sec:estimate}, the intrinsic rewards are valid only when the probability of the selected token lies out of top-$k$ candidates. This raises the question of whether this subset is adequate for curiosity exploration. To investigate, we expand the frequency of intrinsic rewards from top-$1$ ($\approx20\%$) to 40\%, 60\%, 80\% and 100\% of the length of trajectory, selected at random. 

Figure~\ref{fig:ratio} shows the impact on output diversity as the frequency of intrinsic rewards increases. For the TL;DR summarization task, increasing the ratio to 60\% leads to a modest improvement in output diversity, with a 3\% average increase. However, when expanding to cover 100\%, output diversity remains similar to the 20\% case, but alignment quality declines from 0.95 to 0.88. A similar trend is observed on the UltraFeedback dataset: increasing the frequency of intrinsic rewards provides comparable output diversity to the top-$1$ setting, yet significantly reduces alignment quality. These findings suggest that focusing on tokens beyond the top-1 probability is sufficient to identify the contexts in which curiosity-driven exploration is beneficial.

\begin{figure}[t]
    \centering
    \includegraphics[trim={0pt 30pt 0pt 0}, width=0.45\textwidth]{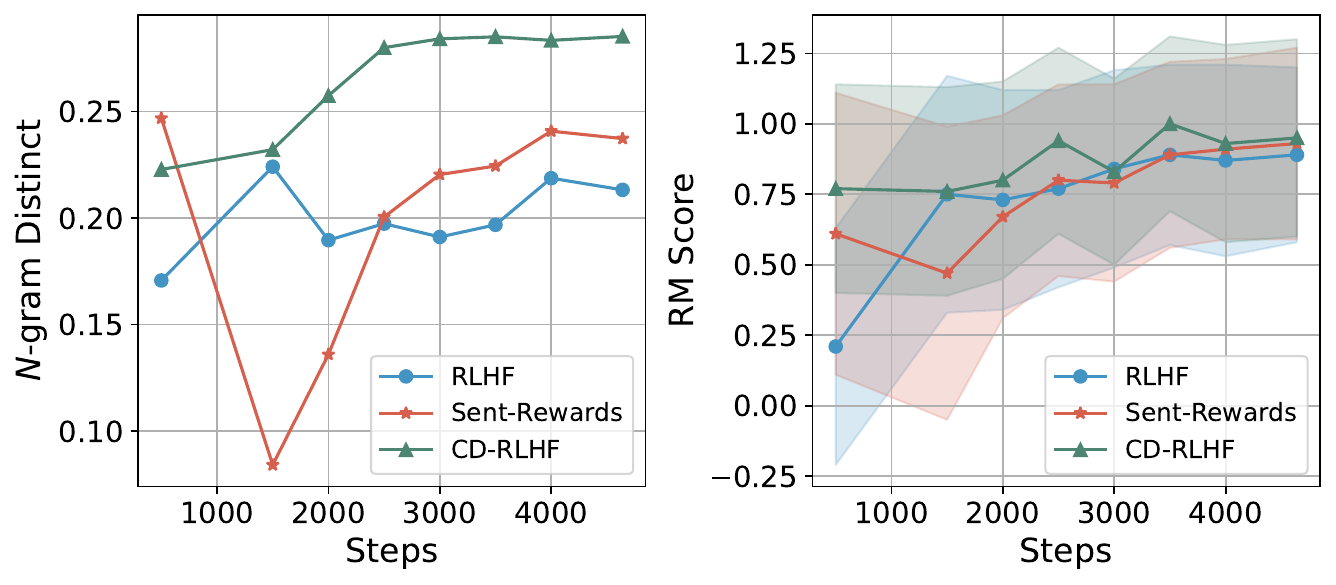}
    \caption{$N$-gram distinct (diversity) and RM score (alignment quality) curves during training for RLHF, Sent-Rewards, and CD-RLHF methods.}
    \label{fig:curve}
    \vspace{-1.5em}
\end{figure}
\paragraph{Training curves of diversity and alignment quality} We compare diversity and alignment quality changes during training for RLHF, CD-RLHF, and Sent-Rewards on the TL;DR task using Gemma-2B model. We use $N$-gram distinct to measure diversity and the RM score to assess alignment quality, with results shown in Figure~\ref{fig:curve}.

For diversity, we observe that RLHF maintains a relatively low diversity level throughout training, after an initial decrease from the SFT model. Both CD-RLHF and Sent-Rewards positively impact diversity, though CD-RLHF consistently improves diversity throughout training, while Sent-Rewards experiences degradation until around 1500 steps.

Regarding alignment quality, all methods show positive effects during training. Notably, CD-RLHF converges faster than the other two, reaching comparable performance at step 2500 rather than at step 4640 (the end of training). Additionally, CD-RLHF displays a more stable alignment quality curve than Sent-Rewards.

\begin{table*}[t]
\centering
\scalebox{0.65}{
\begin{tabular}{l|l|ccc|cccc}
\toprule
\textbf{} & \textbf{} & \multicolumn{3}{c|}{\textbf{MT-Bench Scores $\uparrow$}} & \multicolumn{4}{c}{\textbf{Diversity Metrics}} \\
\midrule
\textbf{Stage} & \textbf{Model} & \textbf{Turn 1} & \textbf{Turn 2} & \textbf{Overall} & \textbf{Diversity $\uparrow$} & \textbf{EAD $\uparrow$} & \textbf{SelfBLEU $\downarrow$} & \textbf{SentBERT $\downarrow$} \\
\midrule
RLHF & \multirow{2}{*}{Gemma-2B} & 6.26 & \textbf{4.45} & 5.35 & 0.1076 & 0.6160 & 0.4383 & 0.7592 \\
CD-RLHF &  & \textbf{6.91} & \textbf{4.45} & \textbf{5.68} & \textbf{0.1123} & \textbf{0.6292} & \textbf{0.3961} & \textbf{0.7315} \\
\midrule
RLHF & \multirow{2}{*}{Gemma-7B} & 6.36 & 5.15 & 5.75 & 0.1173 & 0.5850 & 0.4851 & 0.7854 \\
CD-RLHF &  & \textbf{6.46} & \textbf{5.46} & \textbf{5.96} & \textbf{0.1297} & \textbf{0.6051} & \textbf{0.4623} & \textbf{0.7617} \\
\midrule
RLHF & \multirow{2}{*}{Llama-3.2-1B} & 4.33 & 3.10 & 3.71 & 0.0818 & 0.5356 & 0.4699 & 0.7475 \\
CD-RLHF &  & \textbf{4.78} & \textbf{3.57} & \textbf{4.18} & \textbf{0.0895} & \textbf{0.5629} & \textbf{0.3919} & \textbf{0.7282} \\
\midrule
RLHF & \multirow{2}{*}{Llama-3.2-3B} & 6.47 & \textbf{5.47} & 5.98 & 0.0939 & 0.5878 & 0.4489 & 0.7797 \\
CD-RLHF &  & \textbf{6.71} & 5.45 & \textbf{6.08} & \textbf{0.1133} & \textbf{0.6213} & \textbf{0.3962} & \textbf{0.7529} \\
\bottomrule
\end{tabular}
}
\caption{Results on MT-Bench with MT-Bench scores estimate quality of response, and diversity metrics estimate output diversity. All of the models are trained on the UltraFeedback dataset.}
\label{tab:mtbench}
\vspace{-4mm}
\end{table*}
\subsection{Out-of-Distribution Analysis} \label{sec:ood_benchmark}
In this section, we explore whether the increased output diversity achieved with CD-RLHF impacts generalization capability or potentially enhances it. We use the widely adopted MT-Bench~\citep{mt-bench} as an out-of-distribution (OOD) benchmark to assess general model ability. Since the UltraFeedback dataset is not tailored to a specific task, we evaluate models trained with UltraFeedback on this benchmark. Our quality evaluation metrics include: (1) the absolute scores assigned by GPT-4, and (2) the win rate between responses generated by the RLHF baseline and CD-RLHF.

\begin{figure}[t]
    \centering
    \includegraphics[width=0.33\textwidth]{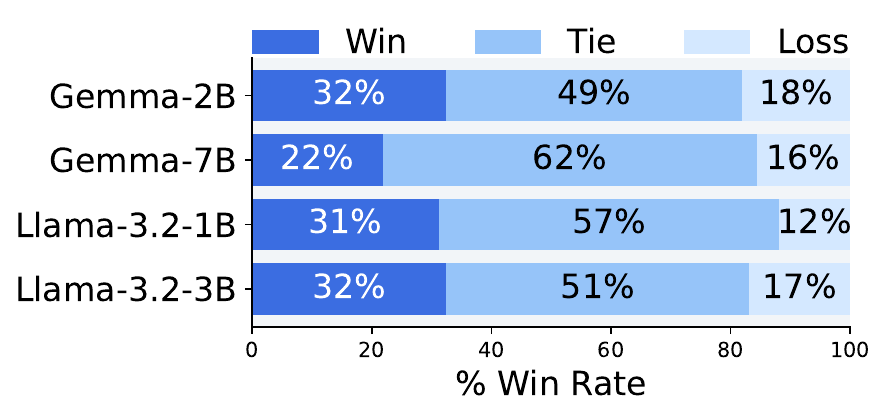}
    \vspace{-1em}
    \caption{Win rates of CD-RLHF against vanilla RLHF on MT-Bench, estimated by GPT-4.}
    \label{fig:mtbench_winrate}
    \vspace{-1.7em}
\end{figure}

In Table~\ref{tab:mtbench}, we present the absolute scores assigned by GPT-4, showing that CD-RLHF outperforms the RLHF baseline for all three models. Notably, although the RM scores results of RLHF and CD-RLHF on Llama-3.2-1B and Gemma-7B models are comparable, CD-RLHF demonstrates superior alignment quality when evaluated by GPT-4. Win rates estimated by GPT-4 are depicted in Figure~\ref{fig:mtbench_winrate}, indicate that CD-RLHF achieves win rates of 32.5\%, 21.875\%, 31\%, and 32.5\% over RLHF on the Gemma-2B, Gemma-7B, Llama-3.2-1B, and Llama-3.2-3B models, respectively.

Apart from the evaluation of quality, we sample 10 answers for each question to estimate the diversity on this OOD benchmark. Since some of questions belong to the categories which sampling temperatures are set to 0.0, we exclude them in this evaluation. Only reserving the questions from writing, roleplay, stem, and humanities categories, with temperature set to 0.7, 0.7, 0.1, and 0.1, respectively. The results are shown in Table~\ref{tab:mtbench}, CD-RLHF consistently exhibits improvements on output diversity, both from syntactic and semantic perspectives evaluated by related metrics. 

\begin{figure}[t]
    \centering
    \includegraphics[trim={0pt 30pt 0pt 0}, width=0.37\textwidth]{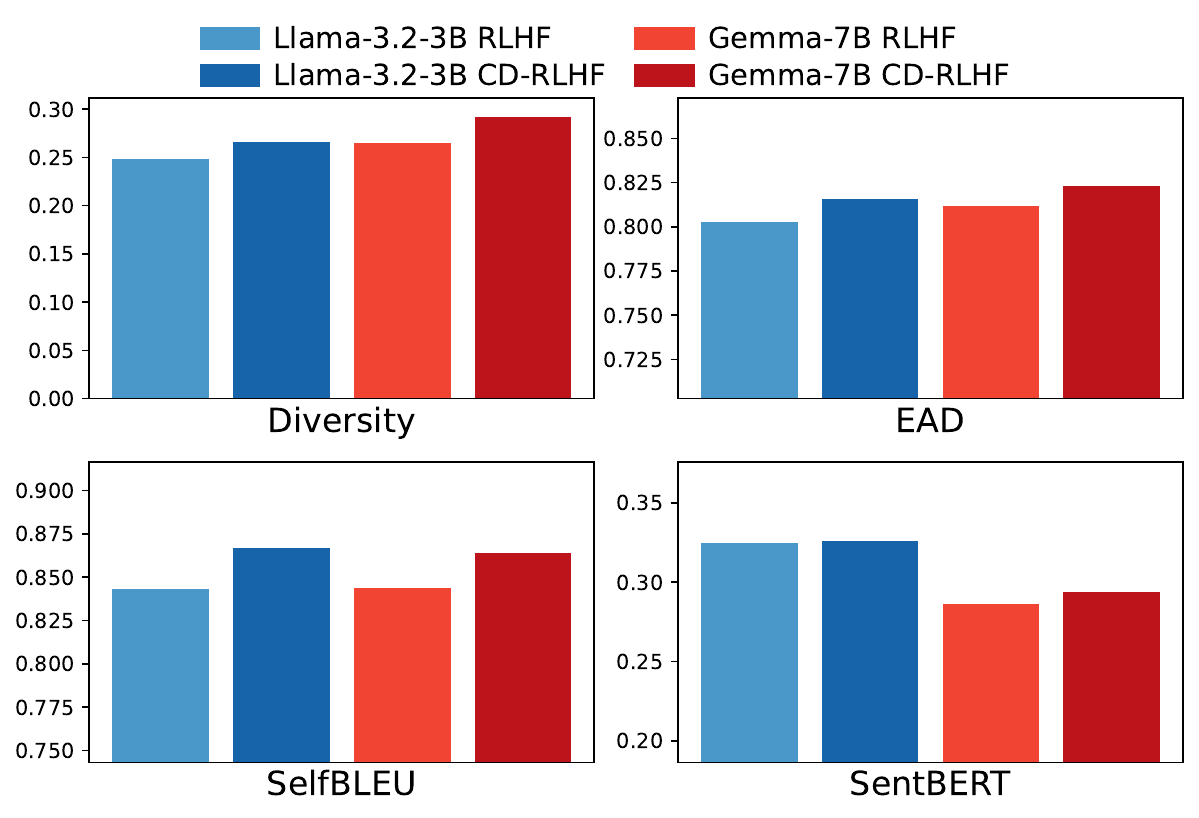}
    \caption{Output diversity of RLHF and CD-RLHF on story writing task. For a better illustration, we use 1.0 - SelfBLEU and 1.0 - SentBERT as metric to maintain a similar trend with other metrics. The higher the better.}
    \label{fig:story}
    \vspace{-1.7em}
\end{figure}

\subsection{Extended Experiments: Story Generation} \label{sec:downstream}
In this section, we assess the output diversity of CD-RLHF in a downstream application: story writing task. We use the validation set of ROC story dataset~\cite{rocstories}, covering 1817 stories. The models involved in this test are the Llama-3.2-3B and Gemma-7B trained on the UltraFeedback dataset, trained with RLHF and CD-RLHF. We prompt the model to generate a story with a start sentence provided by the ROC dataset. The generated story is limited to 200 words. We use the configuration \texttt{temperature=1.0}, \texttt{top\_p=0.9}, and \texttt{top\_k=50}. For evaluation, we use Diveristy, EAD, SelfBLEU, and SentBERT as metrics to assess the output diversity across these stories. The evaluation results are presented in Figure~\ref{fig:story}. On this creative writing task, CD-RLHF shows more diverse outputs than RLHF evidenced by these metrics from both syntactic and semantic. 
\section{Conclusion}
In this paper, we introduce a novel framework, CD-RLHF, which encourages the policy to explore ``novel'' context with curiosity, combining with extrinsic reward to optimize both output diversity and alignment quality. Experiments on the TL;DR summarization and Ultrafeedback instruction-following tasks demonstrate CD-RLHF’s effectiveness in enhancing output diversity while maintaining alignment quality. Additionally, results on an out-of-distribution benchmark highlight the framework's strong generalization capabilities and the benefits of increased output diversity.
\section*{Limitations}
While our work demonstrates the benefits of incorporating a curiosity-driven approach into RLHF to enhance output diversity, CD-RLHF has certain limitations. First, we observed that the intrinsic reward scale is significantly larger than the extrinsic reward. To address this, we set the weight parameter $\eta$ for intrinsic rewards to a small value. However, designing a more suitable ICM that generates intrinsic rewards with a distribution closer to extrinsic rewards could yield better performance. Second, while CD-RLHF mitigates the trade-off between output diversity and alignment quality, this trade-off still persists during the RLHF stage. Finally, although CD-RLHF enhances output diversity without compromising alignment quality, it generally underperforms SFTed models in terms of diversity. In the future, it is promising to further bridge this gap to achieve both the output diversity of SFT and the alignment quality of RLHF.

\bibliography{custom}

\appendix
\clearpage

\begin{table*}[t]
\centering
\scalebox{0.75}{
\begin{tabular}{l|cccccc}
\toprule
\textbf{Dataset}              & \textbf{\begin{tabular}[c]{@{}c@{}}Num. of\\ Comparisons\end{tabular}} & \textbf{\begin{tabular}[c]{@{}c@{}}Num. of\\ Train Samples\end{tabular}} & \textbf{\begin{tabular}[c]{@{}c@{}}Num. of \\ Test Samples\end{tabular}} & \textbf{\begin{tabular}[c]{@{}c@{}}Avg. Tokens\\ in Prompt\end{tabular}} & \textbf{\begin{tabular}[c]{@{}c@{}}Avg. Tokens\\ in Chosen\end{tabular}} & \textbf{\begin{tabular}[c]{@{}c@{}}Avg. Tokens\\ in Rejected\end{tabular}} \\
\midrule
OpenAI Summarization & 179k           & 92.9k            & 86.1k           & 325                   & 35                    & 33                      \\
UltraFeedback        & 63.1k          & 61.1k            & 2k              & 156                    & 282                   & 246                     \\
\bottomrule
\end{tabular}}
\caption{Statistics of datasets involved in experiments. The number of tokens are calculated with Gemma-2B tokenizer.}
\label{tab:statistic}
\end{table*}
\section{Experimental Details} \label{app:training_details}

\subsection{Datasets}

\paragraph{TL;DR Summarization} In this task, the policy is asked to generate summarizations for Reddit posts. This dataset consists of 93k human-annotated preference pairs and 86k pairs for validation. The trainable pairs are derived from the Reddit TL;DR~\citep{reddit_tl;dr} dataset. Additionally, a portion of the validation pairs is sourced from the CNN Daily Mails, which serves as the test set for out-of-distribution generalization. 

\paragraph{UltraFeedback Instruction Following} In this task, the policy needs to complete the given task, which covers various aspects. This dataset is produced with LLMs assist, the completions are generated with different LLM, and scored by GPT-4 on instruction-following, helpfulness, honesty, and truthfulness aspects. We used the binarized version of UltraFeedback, which use the completion with highest overall score as chosen, and one of the remaining 3 at random as the rejected one. This dataset contains 61.1k instances for training and 2k for evaluation. 

We present the data statistics in Table~\ref{tab:statistic}. 

\subsection{Training Details}
\begin{table*}[!t]
\begin{center}
\scriptsize
\begin{tabular}{c|c|cccc}
\toprule
& Hyper-Parameter &  Gemma-2B & Gemma-7B & Llama-3.2-1B & Llama-3.2-3B \\
\midrule
\multirow{5}{*}{SFT}  & Batch size                       & \begin{tabular}[c]{@{}c@{}}512 for TL;DR\\ 128 for UltraFeedback\end{tabular} & 128 & 256  & 256  \\
                      & Epochs                           & 3    & 1 & 3 & 3  \\
                      & Learning rate                    & 5e-5  & 2e-5 & 1e-4 & 5e-5    \\
                      & LR scheduler                     & cosine          & cosine        & cosine  & cosine                  \\
                      & Warmup ratio                     & 0.1               & 0.1            & 0.1    & 0.1    \\
\midrule
\multirow{5}{*}{RM}   & Batch size                       & 64                      & 128   & -  & 64        \\
                      & Epochs          & 1         & 1    & -  & 1   \\
                      & Learning rate                    & \begin{tabular}[c]{@{}c@{}}1e-5 for TL;DR\\ 2e-5 for UltraFeedback\end{tabular} & 1e-6 & - & 1e-5                    \\
                      & LR scheduler                     & cosine         & cosine                                                                & -   & cosine                  \\
                      & Warmup ratio                     & 0.05          & 0.05                                                                 & -    & 0.05                     \\
\midrule
\multirow{16}{*}{PPO} & Batch size                       & 256              & 256    & 256  & 256    \\
                      & Policy learning rate             & \begin{tabular}[c]{@{}c@{}}8e-6 for TL;DR\\ 1e-5 for UltraFeedback\end{tabular} & - & 5e-6          & \begin{tabular}[c]{@{}c@{}}5e-6 for TL;DR\\ 8e-6 for UltraFeedback\end{tabular}           \\
                      & Critic learning rate             & 1e-5      & -                                                              & 1.5e-5        & 1e-5                    \\
                      & Epochs                           & 1          & 1                                                                 & 1       & 1                       \\
                      & PPO epochs                       & 1         & 1                                                                   & 1      & 1                       \\
                      & Rollout                          & 1          & 1                                                                  & 1      & 1                       \\
                      & Clip ratio                       & 0.2        & 0.2                                                                 & 0.2       & 0.2                     \\
                      & $\lambda$ in GAE                 & 0.95        & 0.95                                                                  & 0.95     & 0.95                    \\
                      & $\gamma$ in GAE                  & 1           & 1                                                                 & 1        & 1                       \\
                      & KL coefficient $\beta$                   & 0.05     & 0.05                                                                   & 0.05      & 0.1                   \\
                      & Max prompt length                & 512          & 512                                                                 & 512     & 512                     \\
                      & Max response length              & 512          & 512                                                                 & 512     & 512                     \\
                      & Warmup ratio                     & 0.1          &  0.1  & 0.05 & \begin{tabular}[c]{@{}c@{}}0.05 for TL;DR\\ 0.1 for UltraFeedback\end{tabular}   \\
                      & Temperature  & 0.8                                     & 0.8                  & 0.8    & 0.8 \\
                      & Top-p        & 1.0                                        & 1.0               & 1.0  & 1.0 \\
                      & Top-k        & 50                                    & 50             & 50 & 50 \\
                      & $\eta$ & 0.04 & 0.04 & \begin{tabular}[c]{@{}c@{}}0.06 for TL;DR\\ 0.04 for UltraFeedback\end{tabular} & 0.08 \\
\bottomrule
\end{tabular}
\end{center}
\caption{Hyper-parameters for training Gemma-2B, Gemma-7B, Llama-3.2-1B, and Llama-3.2-3B models for all stages.}
\label{tab:hyperparams}
\end{table*}

Following the procedure of previous work~\citep{instrutGPT}, the SFT model and reward model are fine-tuned on the same dataset with RL fine-tuned model to mitigate the distribution gap. 

\paragraph{SFT Training} We split 20\% of the training instances to conduct the supervised fine-tuning. Since the fine-tuning conducted on the preference dataset, we use the prompt and chosen completion as the instruction data. For TL;DR summarization task, we wrap the instruction with the same format as~\citet{summarize_feedback}: \texttt{POST\escape{n}Subreddit: r/[SUBREDDIT]\escape{n}[PROMPT]\escape{n}TL;DR:[CHOSEN]}. For UltraFeedback instruction following task, we use the ChatML format: \texttt{<|im\char`_start|>user\escape{n}[PROMPT]\escape{n}<|im\char`_end|>\escape{n} <|im\char`_start|>assistant\escape{n}[CHOSEN]\escape{n} <|im\char`_end|>\escape{n}}.

\paragraph{Reward Modelling} We use 40\% of the training instances in this stage. The reward model is initialized using the fine-tuned SFT model.

\paragraph{PPO Training} The remained 40\% of the data is utilized to train the policy model. We initialize the policy and reference model with fine-tuned SFT model, the critic model with reward model. When training Llama-3.2-1B, we use the same reward model as Llama-3.2-3B. We enabled LoRA~\citep{lora} when training 7B model with PPO, and the learning rates of LoRA module in actor and critic model are set to 5e-4, LoRA dimensions are 16. We only insert LoRA into the linear layer of self-attention mechanism. 

When implementing the CD-RLHF, the intrinsic reward whitening operation is used to stable the training procedure, which mitigate the scalar gap across samples.

Table~\ref{tab:hyperparams} lists the hyperparameters used in all stages. The hidden size of encode in ICM is 2 times of hidden size of actor model, and hidden size of ICM is the same as intermediate size of actor model. The experiments are conducted on machines with 8$\times$V100 GPUs or machines with 8$\times$A100 GPUs.

\begin{table}[]
\centering
\scalebox{0.8}{
\begin{tabular}{l|cc}
\toprule
\textbf{Model}        & \textbf{SelfBLEU} & \textbf{SentBERT} \\
\midrule
\multicolumn{3}{c}{TL;DR}                            \\
\midrule
Gemma-2B     & 0.5               & 0.0               \\
Gemma-7B     & 0.5               & 0.5               \\
Llama-3.2-1B & 0.5               & 0.5               \\
Llama-3.2-3B & 1.0               & 1.0               \\
\midrule
\multicolumn{3}{c}{UltraFeedback}                    \\
\midrule
Gemma-2B     & 1.0               & 1.0               \\
Gemma-7B     & 1.0               & 1.0               \\
Llama-3.2-1B & 0.5               & 0.5               \\
Llama-3.2-3B & 0.5               & 0.5               \\
\bottomrule
\end{tabular}}
\caption{Hyper-parameters for training Sent-Rewards method~\cite{curiosity-red-teaming} on Gemma-2B, Gemma-7B, Llama-3.2-1B, and Llama-3.2-3B models.}
\label{tab:sentrewards_params}
\end{table}
For implementing Sent-Rewards, we adjust the weights of SelfBLEU, SentBERT, and entropy rewards in the reward score. In all experiments, we set the entropy weight to 0.01 and apply the entropy penalty as token-level rewards, similar to KL-divergence. The specific weights for SelfBLEU and SentBERT in each experiment are listed in Table~\ref{tab:sentrewards_params}.

\subsection{Evaluation Details}
We evaluate the diversity with the PerInput format. The test set is define as $\mathcal{D}_{\mathrm{test}}$, and $\mathcal{D}_i$ is the $i$-th subset of $\mathcal{D}_\mathrm{test}$, with the completions in $\mathcal{D}_i$ have the same instruction. The details of each metrics are as follows:
\begin{equation*}
\begin{aligned}
\text{rep-}n &= 100 \times(1.0-\frac{N_n}{C_n}) \\
N\mathrm{-gram\ Distinct}_{\mathcal{D}^i}&=\prod^N_{n=1}(1.0 - \frac{\text{rep-}n}{100}) \\
\end{aligned}
\end{equation*}
where $N_n$ is the number of distinct tokens of $n$-gram, and $C_n$ is the total number of tokens of $n$-gram.

\begin{equation*}
    \mathrm{EAD}_{\mathcal{D}^i}=\frac{1}{N}\sum^N_{n=1} \frac{N_n}{V [1 - (\frac{V-1}{V})]^C_n}
\end{equation*}
where $V$ is the vocabulary size.

\begin{equation*}
\begin{aligned}
    \mathrm{SelfBLEU}_{\mathcal{D}^i} &= \frac{1}{N}\sum_{n=1}^{N} \mathrm{SelfBLEU}_{n}(\mathcal{D}^i)
\end{aligned}
\end{equation*}
where $\mathrm{SelfBLEU}_{n}$ is the SelfBLEU calculated with $n$-gram.

\begin{equation*}
    \mathrm{SentBERT}_{\mathcal{D}^i} = \frac{1}{|\mathcal{D}^i|}\sum_{x_k, x_j \in \mathcal{D}^i} \frac{g(x_k)\cdot g(x_j)}{\|g(x_k)\|_2\|g(x_j)\|_2}
\end{equation*}
where $g(\cdot)$ is the embedding model.\footnote{\url{https://huggingface.co/sentence-transformers/all-MiniLM-L6-v2}}

The final results are averaged across all samples for $\mathcal{D}^i \in \mathcal{D}_{\mathrm{test}}$.

For GPT-4 evaluation, we utilize the \texttt{gpt-4o-05-13} to rank the set based on diversity given an instance and two sets of completions, where each set contains 10 completions. We sampled 50 instances from each dataset for evaluation.

For human evaluation, the human annotators are asked to identify which set of completions is more diverse than the other. Unlike the GPT-4 evaluation, the annotators are given an instance and two sets of completions, each containing 5 completions. We sampled 25 instances from each dataset for evaluation.



\section{Comparison Evaluation Details}
\subsection{GPT-4 Evaluation Prompt}
For each evaluation, GPT-4 is employed to assess the diversity of two sets of 10 completions generated by different models, serving as a proxy for human evaluation. All experiments use \texttt{gpt-4o-05-13}. To ensure fairness, the order of sets is randomized across all experiments.

\newtcolorbox{promptbox}[2][Prompt]{
arc=5pt, 
boxrule=0.5pt,
fonttitle=\bfseries,
title=#1, 
before upper={\small}, fontupper=\fontfamily{ptm}\selectfont,
}

\begin{promptbox}[\centering GPT-4 Evaluation Prompt for Ranking Diversity Score]{}
    \vspace{0.2em} 
    Given two sets of responses, your task is to identify which set is more diverse compared to the other. The diversity evaluation should assess the variation among the proposed responses. The more similar the responses within a set, the lower the diversity. And also, you need to identify which set of responses is more related to the prompt:
    
    Prompt: \verb|<instances>|
    
    Set 0: \verb|<completion set 1>|
    
    Set 1: \verb|<completion set 2>|
    
    Your identification (starting with reason, and end with your choice, using ``MY CHOICE: 0/1/2'' to mark the choice, where 0 for Set 0, 1 for Set 1, and 2 for equally diverse):
\end{promptbox}

\begin{promptbox}[\centering GPT-4 Evaluation Prompt for Evaluating Alignemnt Quality on TL;DR]{}
    \vspace{0.2em} 
    You will be given two summaries written for an article. Your task is to pick the better one between them, based on the four criteria. Please make sure you read and understand these instructions carefully.
    
    Relevance - selection of important content from the source. The summary should include only important information from the source document. Annotators were instructed to penalize summaries which contained redundancies and excess information.
    
    Coherence - the collective quality of all sentences. We align this dimension with the DUC quality question of structure and coherence whereby “the summary should be well-structured and well-organized. The summary should not just be a heap of related information, but should build from sentence to a coherent body of information about a topic.”
    
    Consistency - the factual alignment between the summary and the summarized source. A factually consistent summary contains only statements that are entailed by the source document. Annotators were also asked to penalize summaries that contained hallucinated facts.
    
    Fluency - the quality of the summary in terms of grammar, spelling, punctuation, word choice, and sentence structure.
    
    You should output single character to indicate which summary you think is better. ‘A’ stands for Summary A and ‘B’ stands for Summary B. If you think both summaries are equally good, output ‘E’.
    
    Article / Post: \verb|<article / post>|
    
    Summary A: \verb|<summary a>|
    
    Summary B: \verb|<summary b>|
    
    Your Choice (only a single character):
\end{promptbox}

\begin{promptbox}[\centering GPT-4 Evaluation Prompt for Evaluating Alignemnt Quality on UltraFeedback]{}
    \vspace{0.2em} 
    You will be given two responses for an instruction written by different assistants. Your task is to pick the better one between them, based on the three aspects: instruction following, helpfulness, and harmlessness.
    
    You should output single character to indicate which response you think is better. `A' stands for Response A and `B' stands for Response B. If you think both responses are equally good, output ‘E’.
    
    Instruction: \verb|<instruction>|
    
    Response A: \verb|<response a>|
    
    Response B: \verb|<response b>|
    
    Your Choice (only a single character):
\end{promptbox}

\subsection{Human Evaluation Protocol}
Similar to the GPT-4 evaluation, the annotators are given two set of completions generated by two different models, each set contains 5 completions. The human annotators are asked to select the preferred response based on the diversity related criteria:
\begin{enumerate}[leftmargin=*]
    \item Structure Variety: this considers whether these completions have flexible and diverse structures, adapted to the content.
    \item Paraphrasing Variety: this assesses whether these completions rephrase most of the content in unique and varied ways.
\end{enumerate}
The annotators need to choose the more diverse set based on these criteria.

\section{Additional Experiments Results}
\subsection{Alignment Quality Evaluated by GPT-4}
\begin{table}[t]
\centering
\scalebox{0.8}{
\begin{tabular}{l|ccc}
\toprule
\textbf{TL;DR}         & \textbf{Win} & \textbf{Tie} & \textbf{Loss} \\
\midrule
CD-RLHF v.s. SFT        & 39                   & 0            & 11                    \\
CD-RLHF v.s. RLHF      & 27                   & 0            & 23                    \\
\midrule
\textbf{UltraFeedback} & \textbf{Win} & \textbf{Tie} & \textbf{Loss} \\
\midrule
CD-RLHF v.s. SFT     & 31                   & 4            & 15                    \\
CD-RLHF v.s. RLHF      & 21                   & 7            & 22       \\
\bottomrule
\end{tabular}}
\caption{GPT-4 pairwise comparisons of alignment quality demonstrate CD-RLHF's superior win rates over both vanilla RLHF and SFT when evaluating the Llama-3.2-3B model on TL;DR and UltraFeedback benchmarks.}
\label{tab:quality_gpt4}
\end{table}
To thoroughly assess the alignment quality of models trained with SFT, CD-RLHF, and RLHF, we use GPT-4 to perform pairwise comparisons. Specifically, we employ best-of-$N$ sampling to select the response with the highest reward score, where $N$ is set to 10. This experiment is conducted on the Llama-3.2-3B model using both the TL;DR and UltraFeedback datasets. The results, presented in Table~\ref{tab:quality_gpt4}, reveal that CD-RLHF maintains alignment quality comparable to RLHF, while significantly outperforming models trained with SFT.

\subsection{Experimental Results on Top-k}
\begin{table*}[t]
\centering
\begin{tabular}{l|cccc|c}
\toprule
\textbf{} & Diversity$\uparrow$ & EAD$\uparrow$ & SelfBLEU$\downarrow$ & SentBERT$\downarrow$ & RM Score$\uparrow$ \\
\midrule
RLHF & 0.2132 & 0.7347 & 0.3367 & 0.7024 & 0.90 \\
\midrule
CD-RLHF top-$1$ & \textbf{0.2839} & \textbf{0.7793} & \textbf{0.2590} & \textbf{0.6720} & \textbf{0.95}\\
CD-RLHF top-$3$ & 0.2690 & 0.7747 & 0.2669 & 0.6771 & 0.94 \\
CD-RLHF top-$10$ & 0.2613 & 0.7670 & 0.2834 & 0.6833 & \textbf{0.95} \\
\bottomrule
\end{tabular}
\caption{Results on TL;DR dataset with different top-$k$ settings for CD-RLHF.}
\label{tab:topk}
\end{table*}
In Section~\ref{sec:estimate}, the intrinsic reward is valid only when the probability of the selected token lies out of top-$k$. The $k$ is set to 1 in our experiments. In this section, we set $k$ to 3 and 10 for further analysis. Under these setting, the number of intrinsic rewards decreases with the $k$ increases. These experiments are conducted on TL;DR dataset with Gemma-2B model, and the results are shown in Table~\ref{tab:topk}. We observe that with less intrinsic rewards (higher $k$), the output diversity deceases while maintaining the alignment quality.

\section{Algorithm of CD-RLHF}
\begin{algorithm*}
  \centering
\begin{algorithmic}[1]
\caption{Framework of Curiosity-Driven RLHF.}
\label{alg:algorithm1}
\Require the train dataset $\mathcal{D}=\{x_0,x_1,\dots,x_n\}$
\Require Policy model: $\pi_{\mathrm{policy}}$;Reference model: $\pi_{\mathrm{ref}}$; Critic model: $\pi_{\mathrm{critic}}$; Reward model: $\pi_{\mathrm{rm}}$; ICM model: $\pi_{\mathrm{ICM}}$
\For {$x$ \text{in} $\mathcal{D}$}
\State Make experience using policy model $y := \pi_{\mathrm{policy}}(x)$;
\State Get value $V(s_t) := \pi_{\mathrm{critic}}(x,s_t)$ at every time step $t\in[0,|y|)$;    
\State Get reward score at current experience $R := \pi_{\mathrm{rm}}(x,y)$;
\State Compute KL as token-level reward $-\beta D_{\mathrm{KL}}(\pi_{\mathrm{policy}}(a_t)\|\pi_{\mathrm{ref}}(a_t))$
\State Obtain extrinsic rewards $r^{(e)}=R-\beta D_{\mathrm{KL}}(\pi_{\mathrm{policy}}(\cdot)\|\pi_{\mathrm{ref}}(\cdot))$
\State Compute intrinsic rewards $r^{(i)}$ following demonstration in Section~\ref{sec:estimate}
\State Combine intrinsic and extrinsic rewards: $r=r^{(e)}+\eta\cdot r^{(i)}$
\State Obtain advantage function $A_t$ and Q-value function $Q_t$ with GAE$(V(s_t), r)$
\State Optimize policy with $\mathcal{L}_{\text{ppo}}=\hat{\mathbb{E}} \left[\min \left( \frac{\pi_\text{policy} (a_t \mid s_t)}{\pi_{\text{ref}}(a_t \mid s_t)} A_t, \mathrm{clip}( \frac{\pi_\text{policy} (a_t \mid s_t)}{\pi_{\text{ref}}(a_t \mid s_t)}, 1 - \epsilon, 1 + \epsilon) A_t\right)\right]$
\State Optimize critic with $\mathcal{L}_{\mathrm{critic}} = \hat{\mathbb{E}} \left[\|V(s_t) - Q_t\|^2\right]$
\State Optimize ICM with $\mathcal{L}_{\text{ICM}}=\frac{1}{2}\|\hat \phi(s_{t+1}) - \phi(s_{t+1})\|_2^2$
\EndFor
\end{algorithmic}
\end{algorithm*}
In Algorithm~\ref{alg:algorithm1}, we present the procedure of CD-RLHF using PPO.

\section{Case Study}
To better understand the effectiveness of CD-RLHF and its impact on generated completions, we present examples from the TL;DR summarization task and UltraFeedback instruction-following task, using 2B models. For each case, we display the first 5 completions out of 10.

In Table~\ref{tab:appendix_tldr_examples}, we show a case involving a student considering pursuing a Master’s degree but concerned about financial constraints. In the RLHF completions, keywords like ``financially independent'', ``no debt'', ``living stipends'', and ``tuition waivers'' appear frequently across responses. CD-RLHF completions, while including these same keywords, also introduce additional details about the author's identity. For instance, response 3 specifies a tuition fee of \$40K, and response 4 mentions the author’s salary as \$23,000 per year and interest about degree. These CD-RLHF responses maintain core information but show greater diversity compared to RLHF.

Table~\ref{tab:appendix_ultra_examples} demonstrates a case resembling an instruction generation task, where the model generates questions based on a provided ``Background'' and ``Story''. The RLHF completions repeatedly focus on ``demand for steel products'', ``greenhouse gases'', and ``carbon dioxide'', with the term ``increase'' used consistently. In contrast, CD-RLHF completions exhibit more diversity, with keywords like ``global warming'', ``air quality'', ``greenhouse gas'', and ``carbon dioxide''. Notably, response 4 uses ``decrease'' instead of ``increase'', as allowed by the prompt. Additionally, CD-RLHF completions include an "Explanation" to further elaborate on the solution.

\begin{table*}
    \centering
\scalebox{0.85}{
    \begin{tabular}{lp{16.6cm}}
        \toprule
        \textbf{Prompt} & POST
Subreddit: r/personalfinance
**(Let me know if there's a better subreddit for this or if it has been profusely addressed before. I have been browsing this subreddit for a while and have not come across it.)**

I am 23, recently financially independent, with a degree in Linguistics that by itself doesn't do much, but coupled with a Master's could be quite useful. However, getting my masters would put me about \$40,000 in the hole. I just started off, am NOT currently in debt, and don't really want to start my financially independent life by going in debt. 
When I went to undergrad, my dad paid for everything and I didn't have to worry about financial aid, so I have no idea how it works and have never dealt with the system. My question(s) is/are:

1. I know some people get living stipends and tuition wavers for graduate school. How common is this? Have you done so before?
2. Would you recommend going to graduate school? Right now I am making \$23,000 a year and with a degree in speech language pathology could be making significantly more (and it's a field I'm interested in.)

Basically (
TL;DR:  \\
        \midrule
        \multicolumn{2}{c}{RLHF}      \\
        \midrule
        Resp. 1. & 23 year old recently \textcolor{deepgreen}{financially independent}, wondering how to pay for graduate school, if \textcolor{deepgreen}{living stipends} and \textcolor{deepgreen}{tuition wavers} are common and if it would be worth \textcolor{deepgreen}{putting myself in debt}. \\
        \midrule
        Resp. 2. & 23 year old recently \textcolor{deepred}{financially independent} with \textcolor{deepgreen}{no debt} looking for 1) \textcolor{deepgreen}{help with financial aid}/scholarships for graduate school as tuition/housing are prohibitively high and 2) advice on if it would \textcolor{deepgreen}{be beneficial} to go back to school. \\
        \midrule
        Resp. 3. & 23 year old recently \textcolor{deepred}{financially independent}, looking at the possibility of going to graduate school to \textcolor{deepgreen}{build skills and career development}. Know some people get \textcolor{deepred}{living stipends} and \textcolor{deepred}{tuition wavers} for graduate school, how common is that and would you recommend going? \\
        \midrule
        Resp. 4. & 23, recently \textcolor{deepred}{financially independent} with \textcolor{deepred}{no debt}, wanting to know more about \textcolor{deepred}{living stipends} and \textcolor{deepred}{tuition wavers} for graduate school. Would it be worthwhile to attend? \\
        \midrule
        Resp. 5. & 23 year old recently \textcolor{deepred}{financially independent} with a degree but no clue how to proceed with grad school. Know some people get \textcolor{deepred}{living stipends} and \textcolor{deepred}{tuition wavers}; would you recommend going to graduate school? \\
        \midrule
        \multicolumn{2}{c}{CD-RLHF}      \\
        \midrule
        Resp. 1. & 23, recently \textcolor{deepgreen}{financially independent}, wanting to apply for graduate school next fall (\textcolor{deepgreen}{masters}), seeking advice on getting \textcolor{deepgreen}{living stipends} and \textcolor{deepgreen}{tuition wavers}. Would it be worth pursuing? \\
        \midrule
        Resp. 2. & 23/f recently \textcolor{deepred}{financially independent}, recently thinking about \textcolor{deepgreen}{pursuing grad school} to further my career. Do you get \textcolor{deepred}{living stipends} and \textcolor{deepred}{tuition wavers} for grearndschools and would you recommend going? \\
        \midrule
        Resp. 3. & 23, recently \textcolor{deepred}{financially independent}, looking at grad school for \textcolor{deepgreen}{40K} tuition in field that could earn much more monetarily. Want to know if it is possible/worth it for grad school as well as \textcolor{deepgreen}{advice} on grad school process. \\
        \midrule
        Resp. 4. & 23 year old recently \textcolor{deepred}{financially independent} - recently learned about graduate school. Wondering whether anyone has been granted \textcolor{deepred}{living stipends} or \textcolor{deepred}{tuition waivers} - would you recommend going to grad school? Right now making \textcolor{deepgreen}{\$23,000 a year} and \textcolor{deepgreen}{interested in SPEECH/PHARM/PEDIATRIC psychology}. \\
        \midrule
        Resp. 5. & 23 yr old \textcolor{deepgreen}{college alumni} with \textcolor{deepgreen}{degree in linguistics} needing \textcolor{deepgreen}{financial assistance} for graduate school-has anyone gotten \textcolor{deepred}{living stipends} and/or \textcolor{deepred}{tuition wavers} for graduate school? Would it be a wise move financially and is there any info on the process/realities? \\
        \bottomrule
    \end{tabular}}
    \vspace{4mm}
    \caption{Example responses from TL;DR dataset. We mark the keywords that showed the first time with \textcolor{deepgreen}{green}, and \textcolor{deepred}{red} for repeated keywords for better illustration.}
    \label{tab:appendix_tldr_examples}
\end{table*}
\begin{table*}
    \centering
\scalebox{0.80}{
    \begin{tabular}{lp{16.6cm}}
        \toprule
        \textbf{Prompt} & You are given a background paragraph that describes one or more causal or qualitative relationships such as a relationship in economics or a scientific law and a story that makes use of the concepts or the relationship described in the provided paragraph. You need to come up with a question about the story that requires an understanding of the relationship described in the background paragraph. The generated question should not be answerable without both the background and story. Write a question about the story that requires a relationship in the background paragraph to answer. Check whether your question requires both the background paragraph and the story to answer. If your question can be answered by only one of these, try to rewrite your question so that it requires both. Note that "flipping" a word in the question can give rise to a new question where the answer will be different from the original question. Often, "flipping" a word gives rise to flipping the direction of the relationship, for example, from "increase" to "decrease."
One example: 
Story: The AAA auto club does a weekly report on the prices of gas and diesel in various cities. Historically, the prices have be fairly stagnant, but this week their report garnered widespread attention as the prices of gas and diesel plunged from last week in Seattle by \$0.40 each. However, just across the border in Vancouver, the cost of gas went up dramatically by \$0.50.
Solution is here: Which city will have an increase in demand for gas?
Explanation: This is a good question as it requires understanding of the relationship described in both the background and story (the background mentions "a fall in price will increase the quantity demanded" and the story cites the decreasing price of gas in Seattle).

Now, solve this: Background Paragraph: Pollutants also affect the atmosphere through their contribution to global warming. Global warming is an increase in the Earth’s temperature. It is thought to be caused mostly by the increase of greenhouse gases like carbon dioxide. Greenhouse gases can be released by factories that burn fossil fuels. Over the past 20 years, burning fossil fuels has produced about three-quarters of the carbon dioxide from human activity. The rest of the carbon dioxide in the atmosphere is there because of deforestation, or cutting down trees ( Figure below ). Trees absorb carbon dioxide during cellular respiration, so when trees are cut down, they cannot remove carbon dioxide from the air. 
Story: Tokyo recently had a large explosion in one of their largest steel manufacturing plants. The combination of heat, steel, and manufacturing materials released an extraordinary amount of carbon dioxide into the atmosphere. Meanwhile, Osaka has dramatically decreased the amount of carbon dioxide it releases through the installation of many public transportation projects and tree planting initiatives.
Solution:  \\
        \midrule
        \multicolumn{2}{c}{RLHF}      \\
        \midrule
        Resp. 1. & Which city will have an \textcolor{deepgreen}{increase} in \textcolor{deepgreen}{demand for steel products}? \\
        \midrule
        Resp. 2. & Which city will experience an \textcolor{deepred}{increase} in \textcolor{deepred}{demand for steel products} due to the Tokyo explosion? \\
        \midrule
        Resp. 3. & Which city will experience an \textcolor{deepred}{increase} in the \textcolor{deepgreen}{release of greenhouse gases} following the Tokyo explosion? \\
        \midrule
        Resp. 4. & Which city will have an \textcolor{deepred}{increase} in the amount of \textcolor{deepgreen}{carbon dioxide released} into the atmosphere? \\
        \midrule
        Resp. 5. & Which city experienced an \textcolor{deepred}{increase} in the \textcolor{deepred}{release of carbon dioxide} following the Tokyo explosion? \\
        \midrule
        \multicolumn{2}{c}{CD-RLHF}      \\
        \midrule
        Resp. 1. & Which city experienced an \textcolor{deepgreen}{increase} in \textcolor{deepgreen}{global warming} as a result of the Tokyo steel plant explosion?

Explanation: ... \\
        \midrule
        Resp. 2. & Which city experienced an \textcolor{deepred}{increase} \textcolor{deepgreen}{in air quality} due to \textcolor{deepgreen}{the decrease in carbon dioxide emissions}?

Explanation: This question requires understanding of the relationship described in the background paragraph (global warming caused by greenhouse gases) and the story (the accidental explosion in Tokyo released carbon dioxide into the atmosphere, while Osaka decreased their carbon dioxide emissions through public transportation and tree planting initiatives). \\
        \midrule
        Resp. 3. & Which city's \textcolor{deepgreen}{greenhouse gas emission} will \textcolor{deepred}{increase} as a consequence of the Tokyo explosion?
        
Explanation: ... \\
        \midrule
        Resp. 4. & Which city will have a \textcolor{deepgreen}{decrease} in \textcolor{deepgreen}{demand for steel products}?

Explanation: In the given background, it is stated that the excessive release of carbon dioxide from a large-scale manufacturing plant in Tokyo can lead to a decrease in demand for steel products. On the other hand, Osaka has taken initiatives to decrease carbon dioxide emissions through public transportation and tree planting projects, which can potentially increase the demand for steel products. Therefore, this question requires an understanding of the relationship between carbon dioxide emissions and demand for steel products, as well as the context of the given stories. \\
        \midrule
        Resp. 5. & Which city will have an \textcolor{deepred}{increase} in \textcolor{deepred}{carbon dioxide emissions}?

Explanation: ...\\
        \bottomrule
    \end{tabular}}
    \vspace{4mm}
    \caption{Example responses from Ultrafeedback dataset. We omit some explanations in responses. We mark the keywords that showed the first time with \textcolor{deepgreen}{green}, and \textcolor{deepred}{red} for repeated keywords for better illustration.}
    \label{tab:appendix_ultra_examples}
\end{table*}

\end{document}